\documentclass[a4paper,conference]{IEEEtran}

\usepackage{amsmath,amsfonts,amssymb}
\usepackage{multirow}
\usepackage{hyperref}
\usepackage{graphicx}
\usepackage{booktabs}
\usepackage{pifont}
\usepackage{enumitem}
\usepackage[table]{xcolor}
\usepackage{scrextend}
\usepackage{dsfont}
\definecolor{lightblue}{rgb}{0.93,0.95,1.0}

\newcommand{\cmark}{\ding{51}}%
\newcommand{\xmark}{\ding{55}}%

\definecolor{arylideyellow}{rgb}{0.91, 0.84, 0.42}
\newcommand{\mybullet}{\raisebox{1.5pt}{\scriptsize $\blacktriangleright$}}

\ifCLASSINFOpdf
\else
\fi
\begin{document}
%
\title{Multimodal Emotion Recognition with Modality-Pairwise Unsupervised Contrastive Loss}

\author{\IEEEauthorblockN{Riccardo Franceschini}
\IEEEauthorblockA{Eurecat, Centre Tecnològic de Catalunya\\
Cerdanyola del Valles, Spain\\
riccardo.franceschini@eurecat.org}
\and
\IEEEauthorblockN{Enrico Fini, Cigdem Beyan, Alessandro Conti, Federica Arrigoni}
\IEEEauthorblockA{Department of Information Engineering and Computer Science\\
University of Trento, Trento, Italy\\
\{enrico.fini,cigdem.beyan,alessandro.conti-1,federica.arrigoni\}@unitn.it}
\and
\IEEEauthorblockN{Elisa Ricci}
\IEEEauthorblockA{\hspace{3cm} Department of Information Engineering and Computer Science, University of Trento, Trento, Italy\\
Fondazione Bruno Kessler, Trento, Italy\\
e.ricci@unitn.it}}


%
\author{\IEEEauthorblockN{Riccardo Franceschini\IEEEauthorrefmark{1},
Enrico Fini\IEEEauthorrefmark{2},
Cigdem Beyan\IEEEauthorrefmark{2},
Alessandro Conti\IEEEauthorrefmark{2},
Federica Arrigoni\IEEEauthorrefmark{2}, and
Elisa Ricci\IEEEauthorrefmark{2}\IEEEauthorrefmark{3}}
\IEEEauthorblockA{\IEEEauthorrefmark{1}Eurecat, Centre Tecnològic de Catalunya, Cerdanyola del Valles, Spain, riccardo.franceschini@eurecat.org}
\IEEEauthorblockA{\IEEEauthorrefmark{2}Department of Information Engineering and Computer Science,
University of Trento, Trento, Italy\\
\{enrico.fini, cigdem.beyan, alessandro.conti-1, federica.arrigoni, e.ricci\}@unitn.it}
\IEEEauthorblockA{\IEEEauthorrefmark{3}Fondazione Bruno Kessler (FBK), Trento, Italy}
}


\maketitle

\begin{abstract}
Emotion recognition is involved in several real-world applications. With an increase in available modalities, automatic understanding of emotions is being performed more accurately. The success in Multimodal Emotion Recognition (MER), primarily relies on the supervised learning paradigm. However, data annotation is expensive, time-consuming, and as emotion expression and perception depends on several factors (e.g., age, gender, culture) 
obtaining labels with a high reliability is hard. Motivated by these, we focus on unsupervised feature learning for MER. 
We consider discrete emotions, and as modalities text, audio and vision are used.
Our method, as being based on contrastive loss between pairwise modalities, is the first attempt in MER literature. Our end-to-end feature learning approach has several differences (and advantages) compared to existing MER methods: 
i) it is unsupervised, so the learning is lack of data labelling cost; ii) it does not require data spatial augmentation, modality alignment, large number of batch size or epochs; iii) it applies data fusion only at inference; and iv) it does not require backbones pre-trained on emotion recognition task. The experiments on benchmark datasets show that our method outperforms several baseline approaches and unsupervised learning methods applied in MER. 
Particularly, it even surpasses a few supervised MER state-of-the-art.
\end{abstract}


%
\IEEEpeerreviewmaketitle

\section{Introduction}
\label{sec:intro}

Emotion is a key factor driving people's actions and thoughts, and a fundamental part of the human verbal and nonverbal communication. 
Automated emotion recognition is an important aspect of many applications,
including 
social assistive robots \cite{SPRING}, 
smart systems to work in 
customer service \cite{Burkhardt2006}, health-care 
\cite{dhuheir2021}, education 
\cite{Hammoumi2018},
and automated-driving cars 
\cite{Pavan2021}.
However, it is a highly challenging problem due to the complex nature of emotion \emph{expression} and \emph{perception}, which are hard to generalize as being dependent on several factors such as age 
\cite{Demenescu2014}, gender 
\cite{Olderbak2019}, cultural background \cite{Engelmann2013}, and personality traits \cite{Furnes2019}. Furthermore, as humans can express their emotions across various modalities (e.g., language, facial expressions, gestures, and speech), it is essential to effectively model the interactions between these modalities, containing complementary but also (possibly) redundant information \cite{Baltrusaitis2019}.

The majority of works 
mainly concentrated on unimodal learning of emotions \cite{beyan2021modeling,AbdullahAhmadAl20,ShirianTripathiAl21}, i.e., processing a single modality. Although there exist breakthrough achievements by unimodal emotion recognition, due to the aforementioned multimodal nature of emotion expression, such models remain incapable in some circumstances. On the other hand, multimodal emotion recognition (MER) holds the challenges of multimodal machine learning, e.g., representing the data to be able to exploit the complementarity and redundancy of modalities, data translation among modalities, co-learning, modality alignment (e.g., capturing temporal information) and data fusion (see \cite{Baltrusaitis2019} for details).
Like most intelligent systems, the advancements in deep learning have enhanced MER, particularly, by utilizing the abundance of data availability. Studies in this field 
(e.g., \cite{TsaiMaAl20,ZhangLiangqingAl19,WenYouAl21,HoYangAl20}) so far, treat the learning process with the supervised way, thus require an intense labor for annotations. 

This paper addresses the problem of \textbf{\emph{perceived multimodal emotion recognition}} when the emotions are represented as \textbf{\emph{discrete categories}} and, more importantly, we learn the features in an \textbf{\emph{unsupervised fashion}}. Motivated by the fact that contrastive learning has shown accurate and robust performance in many domains 
(e.g., \cite{chen2020simple,Rai_2021_CVPR}), we adapt the contrastive loss function \cite{Luyu2020} to perform pairwise modality feature learning.
To the best of our knowledge, this is the first time contrastive loss is adapted for MER. Our approach learns feature embeddings in an end-to-end fashion (see \cite{DaiCahyawijayaAl21} for the definition), and differs from the prior works in terms of several aspects, which are described as follows. \\ 
\noindent
\textbf{i) Modality exploitation.} Our method leverages different modalities in a contrastive learning framework. Given a data sample represented in terms of multiple modalities, our aim is to push the embeddings of two modalities of the same sequence to be close to each other 
while pulling the embeddings of the same two modalities of different sequences to be apart. 
Note that the sequences that are being pulled apart can be from the same class. But, herein we \emph{do not use the class labels}, thus we only aim to make the representations of the same sequence across modalities similar (as close as possible) to each other. \\
\noindent
\textbf{ii) Data translation $\&$ co-learning.} We contrast the feature embedding of one modality with another modality when both are belonging to the \emph{same data sample}.
This can be seen as an analogy to performing \emph{data translation} and ultimately \emph{co-learning}. Unlike existing contrastive learning approaches 
(e.g., \cite{Haocong2021,Rai_2021_CVPR}),
we do not require data spatial augmentation (e.g., random crops, blurs or color distortions). Also, different from approaches \cite{chen2020simple,Chen2020,MaoLi2021} relying on heavy data augmentations as well as large number of batch sizes and epochs, our method is much more affordable. \\
\noindent
\textbf{iii) Modality alignment.} The outputs of different sensors might have different (but fixed) sample rates. However, this is not valid for text, which makes obtaining word-aligned sequences not so obvious \cite{WenYouAl21}. Still, multimodal data alignment is an imperative step to perform an effective MER for several methods 
(e.g., \cite{Koromilas2021UnsupervisedML,ShenoySardanaAl20}), 
resulting in the real-world application of such methods challenging. In contrast, our method does not require \emph{perfectly aligned modalities}. We considered both aligned samples and a mixture of aligned/misaligned samples in our experiments
(Sec.~\ref{subsec:datasets}). \\
\noindent
\textbf{iv) Data fusion.} 
It is applied here only at inference via the concatenation of learned feature representations. This is different from the MER state-of-the-art (SOTA) applying data fusion \emph{both} in training and testing \cite{RadoiBirhalaAl21,ShenoySardanaAl20,SongCaiAl21,MittalBhattacharyaAl20}. \\
\noindent
\textbf{v) Data labelling.} Our method is free from data labeling cost by being an \emph{unsupervised feature learning} approach. Note that there exist a few number of unsupervised approaches in the same and/or related topics, e.g., speech emotion recognition \cite{Neumann2019,Zhang2021}, facial emotion recognition \cite{Xiao2019}, facial expression intensity estimation \cite{AwiGra2018a}, and multimodal sentiment and emotion analysis \cite{Koromilas2021UnsupervisedML}.
However, our method involves the deep architectures either pre-trained on tasks different from emotion recognition (e.g., action recognition) or \emph{not} pre-trained. This aspect introduces a potential to apply the proposed method to the related downstream tasks, e.g., multimodal sentiment analysis and social interaction analysis, without the need of customization. Some approaches (e.g., \cite{Hu_2018_ECCV,Savchenko2021,Minji2020}), instead, could supply the desired performance (e.g., outperforming the best of all methods of comparison time) if and only if they are pre-trained on large emotion datasets having the \emph{same emotion labels} as in the test set.

\smallskip

To validate the effectiveness of our method, experiments were realized on two multimodal emotion datasets.
Results show that the proposed method outperforms prior unsupervised MER approaches and several baselines. Moreover, despite performing unsupervised feature learning, our method even surpasses some of the fully-supervised MER methods. 
%
To summarize, the main contributions of this study are: (1) presenting a novel unsupervised multimodal feature learning approach, (2) being the first study adapting the contrastive loss for MER, and (3) improving the emotion recognition results compared to unsupervised feature learning MER SOTA.
The code of the proposed method is available at \url{https://github.com/ricfrr/mpuc-mer}.

\begin{figure*}[th!]
    \centering
    \includegraphics[width=0.65\textwidth]{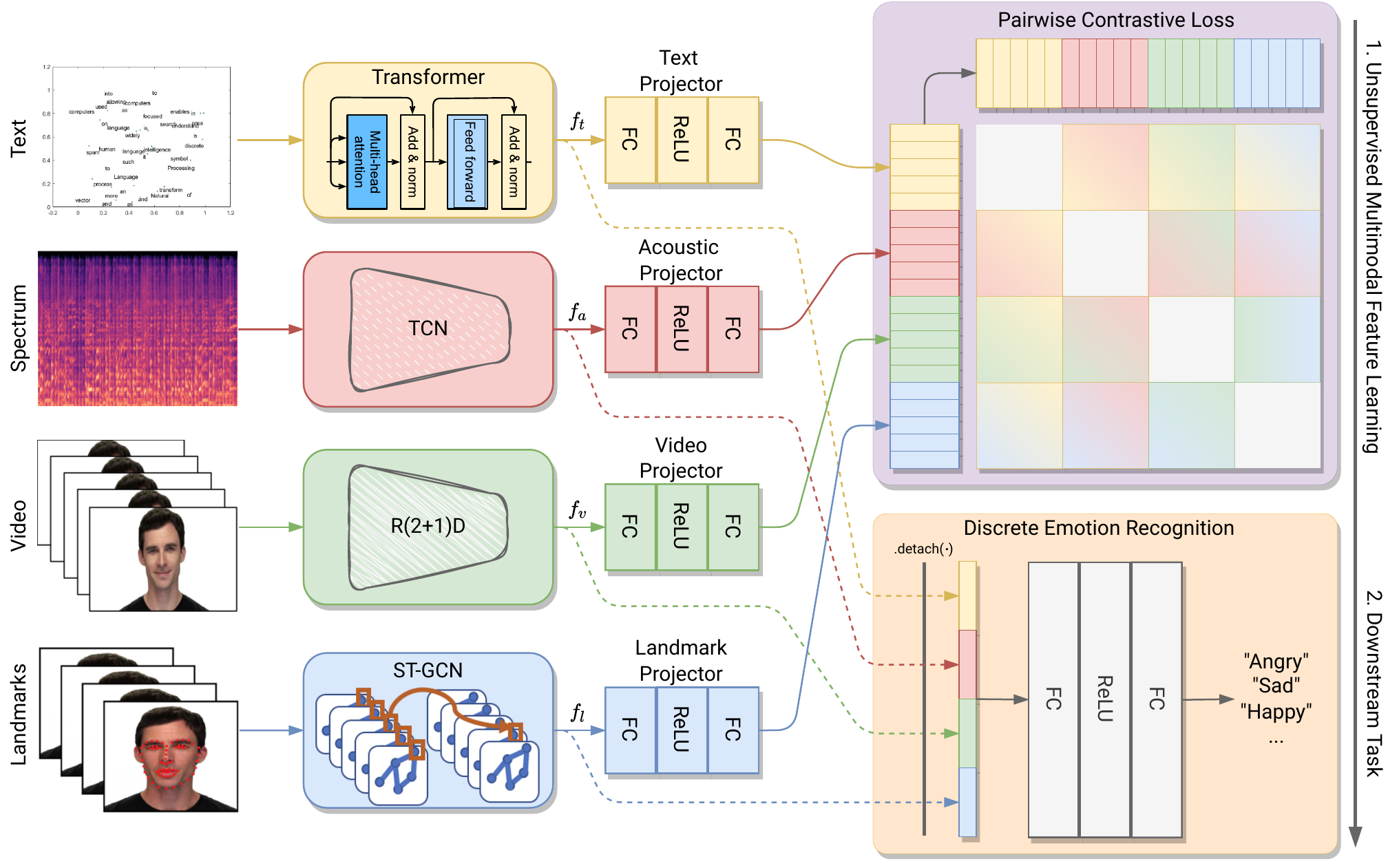}
    \caption{\textbf{Summary of our approach.} We first learn the multimodal features in an unsupervised fashion, then the downstream task (discrete emotion recognition) is performed. We jointly train, each possible pair of modalities' backbone 
    using contrastive loss in order to predict the correct pairings of a batch of 
    training examples. 
    The final loss is the average of all losses calculated. During inference, $f_t, f_a, f_v, f_l$ are extracted before the projection layers (i.e., $fc$+RELU+$fc$) and concatenated, then feed to a linear classifier for emotion recognition.}
    \label{fig:proposedM}
\end{figure*}

\section{Related Work}
\label{sec:relatedwork}

%
Several methods for multimodal emotion recognition (MER) were proposed, as detailed in the recent survey papers: \cite{SpezialettiPlacidiAl20,Sharma2021}. In this section, our summary is regarding \emph{discrete} MER research modeling text, visual and acoustic modalities, as we tested our method on that context. 
Early works adapt classifiers like SVMs, Linear and Logistic Regression 
\cite{Castellano2008,Sikka2013} 
while, by the time bigger datasets were developed, deep learning architectures were also explored. 
For example, \cite{RadoiBirhalaAl21} 
is based on CNNs, and \cite{ShenoySardanaAl20,SongCaiAl21} use RNNs. Some recent studies 
\cite{DelbrouckTitsAl20,TsaiMaAl20,WenYouAl21} 
adopt Transformers. 

Ghaleb et al. \cite{GhalebPopaAl19} apply deep metric learning in which a LSTM component models the variations of the emotions as a function of time. That is different from late fusion of modalities \cite{RadoiBirhalaAl21,SongCaiAl21} or building temporal features to extract global information by assuming that emotions are expressed simultaneously \cite{ShenoySardanaAl20}.
Late fusion is favorably applied by concatenating the learned features of all modalities in \cite{RadoiBirhalaAl21,SongCaiAl21} or with a pairwise scheme in \cite{ShenoySardanaAl20}. Instead, the authors of M3ER \cite{MittalBhattacharyaAl20} propose a data-driven multiplicative fusion method to combine the modalities, which learns to emphasize the more reliable cues and suppresses the others by integrating Canonical Correlation Analysis as a pre-processing step.
Differently, Zadeh et al. \cite{Zadeh2018MultimodalLA} present Graph-MFN, 
which synchronizes the multimodal sequences by storing intra-modality and cross-modality interactions through time with a graph structure. 
Attention mechanism has been exploited by several works as well \cite{BeardDasAl18,ChoiSongAl18,DelbrouckTitsAl20,ChauhanAkhtarAl19,AkhtarChaudanAl20,ZhangLiangqingAl19,HoYangAl20,GhalebNiehuesAl20,DaiCahyawijayaAl21,KhareParthasarathyAl21}.
For example, Dai et al. \cite{DaiCahyawijayaAl21} present MESM that is composed of sparse cross-modal attention mechanism attached to the joint learning of multimodal features.

There are a lot of attempts applying end-to-end learning \cite{RadoiBirhalaAl21,ShenoySardanaAl20,ChangSkarber21,HuynhVan2021}, but only \cite{DaiCahyawijayaAl21} compared a fully end-to-end method (defined as jointly optimizing feature extraction and feature learning stages \cite{DaiCahyawijayaAl21}) with the two-phase pipelines (i.e., feature extraction is independent from multimodal learning). 
Indeed, it is very common in the MER litreature to apply the feature extraction step separately. This is performed on each modality by using either hand-crafted formulations \cite{TiwariRathodAl21,JaratrotkamjornChoksuriwong19,MittalBhattacharyaAl20,ShenoySardanaAl20,RadoiBirhalaAl21,Zadeh2018MultimodalLA}) and/or deep learning architectures \cite{GhalebPopaAl19,ShenoySardanaAl20,RadoiBirhalaAl21}. As example of acoustic features; Log-Mel spectrogram \cite{RadoiBirhalaAl21}, pitch, voiced/unvoiced segmenting features \cite{ShenoySardanaAl20,Zadeh2018MultimodalLA,MittalBhattacharyaAl20}, MFCCs \cite{SongCaiAl21,ShenoySardanaAl20,Zadeh2018MultimodalLA,MittalBhattacharyaAl20}, features extracted from SoundNet 
\cite{GhalebPopaAl19}) can be given.
On the other hand, various backbones such as VGG16 
\cite{SongCaiAl21}, I3D 
\cite{GhalebPopaAl19}, FaceNet 
\cite{GhalebPopaAl19} as well as facial features; facial landmarks and facial action units extracted by OpenFace 
\cite{Zadeh2018MultimodalLA,MittalBhattacharyaAl20} are among the most popular visual features.
For text, 
Glove embeddings \cite{pennington2014} have been frequently utilized \cite{ShenoySardanaAl20,Zadeh2018MultimodalLA,MittalBhattacharyaAl20,DelbrouckTitsAl20,DelbrouckTitsAl20b,WenYouAl21,TsaiMaAl20}, while Transformers are used as the backbone \cite{DelbrouckTitsAl20,DelbrouckTitsAl20b,WenYouAl21,TsaiMaAl20,KhareParthasarathyAl21} or LSTMs are trained with the extracted word embeddings \cite{Zadeh2018MultimodalLA,MittalBhattacharyaAl20}.

Among the aforementioned approaches, \cite{ChoiSongAl18,DelbrouckTitsAl20b} use text and audio, \cite{JaratrotkamjornChoksuriwong19,GhalebNiehuesAl20,GhalebPopaAl19,RadoiBirhalaAl21,SongCaiAl21,ChangSkarber21,TiwariRathodAl21} use video and audio, and all others use text, audio and video together. 
It is worth noting that these techniques are all \textbf{\emph{supervised}}. Recently, Khare et al. \cite{KhareParthasarathyAl21} investigated the usage of large unlabeled multimodal datasets for pre-training a cross-modal transformer, which is then fine-tuned for the emotion recognition task. 
In detail, 
the VoxCeleb dataset \cite{Chung2018VoxCeleb2DS}, composed of 1.1 million videos that are associated to emotions \cite{Albanie2018}, is used to pre-train the multimodal transformer. Then, the decoder layer is removed, and an average pooling and additional fully connected layers are added to fine-tune the model for emotion recognition task.
Unlike \cite{KhareParthasarathyAl21}, we do not rely on auxiliary large-scale datasets to pre-train our model, and both the feature learning and inference are performed on the same datasets, which are much smaller than the VoxCeleb dataset \cite{Chung2018VoxCeleb2DS}. Our learned features are frozen such that we do not apply any fine-tuning as in \cite{KhareParthasarathyAl21}. This is an important difference because some studies \cite{UHAR_BMVC2021,li2021crossclr} have shown that, compared to using frozen features that are learned in an unsupervised fashion, fine-tuning can bring up to 17.5\% improvement for the downstream task. However, following the fine-tuning approach would not keep the feature learning methodology \emph{``entirely unsupervised''}, as it requires the labels of the downstream task.
Moreover, our model is applicable with different modality combinations, whereas text is an anchor modality in \cite{KhareParthasarathyAl21}.

The MER litreature is very limited in terms of \textbf{\emph{fully unsupervised feature learning}} approaches. Very recently, a Convolutional Autoencoder architecture is presented in \cite{Koromilas2021UnsupervisedML}.
Despite being very different from our method in terms of the architecture, \cite{Koromilas2021UnsupervisedML} is still our ``direct competitor'' by having the following common aspects with the proposed method:  
\textit{i)} performing unsupervised feature learning without fine-tuning, \textit{ii)} being independent to the number of modalities and modality combinations, and \textit{iii)} not being task-specific.




\section{Our Approach}
\label{sec:propsoedMethod}

An overview of our approach is given in Fig.~\ref{fig:proposedM}. First, the multimodal features are learned with an unsupervised way (Sec.~\ref{subsec:unsupervised}). Then, the downstream task (discrete emotion recognition) is performed (Sec.~\ref{subsec:emotionRec}).
Sec.~\ref{subsec:features} describes the modalities 
and Sec.~\ref{subsec:training} presents the implementation details.

\subsection{Modalities}
\label{subsec:features}
The modalities and backbones we utilize are described as follows.\\
\noindent
\mybullet\ \textbf{Text.} The word vectors are extracted from transcripts with the Glove word embeddings \cite{pennington2014}, following the procedure in \cite{Zadeh2018MultimodalLA}. As the backbone, we use the Transformer in \cite{Vaswani2017}, which is one of the SOTA architectures of language processing. \\
\noindent
\mybullet\ \textbf{Visual.} We rely on two sources of visual data. One of them is
the \textbf{facial images} extracted by MTCNN face detector \cite{zhang2016joint} (unless faces are supplied by the dataset) from RGB video frames. As the backbone associated to the facial images, the R(2+1)D architecture \cite{Tran2018} pre-trained on Kinetics-400 dataset \cite{Kay2017} is used. The other visual data is the \textbf{facial landmarks} detected by the method in \cite{bulat2017far} (unless it is provided by the dataset used), and the associated backbone is Spatio-Temporal Graph Neural Network (ST-GCN) \cite{Bing2018}. \\ 
\noindent
\mybullet\ \textbf{Acoustic.} Mel-spectograms are extracted with the same procedure and settings in \cite{Tachibana2018,DelbrouckTitsAl20,DelbrouckTitsAl20b} with Librosa Python Library \cite{mcfee2015librosa} using 80 filter banks and by selecting one frame for every 16 frames. The dimension of the mel-spectograms is fixed to 128. We adapt Time Convoluted Network (TCN) \cite{Pariente2020Asteroid} such that it takes mel-spectrograms as the input.

As seen, each modality has its own backbone, which have been chosen as being the SOTA architectures for diverse applications of language, visual and acoustic data processing.

\subsection{Unsupervised Multimodal Feature Learning}
\label{subsec:unsupervised}

The proposed method includes separate multi-layer projection heads onto each backbone defined in Sec.~\ref{subsec:features}. All projection heads have the same structure such that they are composed of fully-connected layers ($fc$), where the first layer is followed by a ReLU activation function ($fc_{1}+ReLU+fc_{2}$).
This structure is motivated by SimCLR \cite{chen2020simple}, which shows that a nonlinear projection head contributes to the performance more than a linear projection head, and its contribution is even more compared to not including any projection layer.

We adapt the CLIP fashion \cite{radford2021learning} training, \emph{without using any labels} of the downstream task (i.e., emotion recognition). 
Given a data sample represented by a sequence of observations in multiple modalities, our aim is to make the embeddings of two modalities of the same sequence (\textit{positives}) close to each other, and make the embeddings of the same two modalities of different sequences (\textit{negatives}) apart from each other. This is repeated for all possible pairs of modalities.
Notice that negative samples might belong to the same class (\textit{i.e.} exhibit the same emotion). However, herein, we assume that the class labels are not available, and we resort to instance discrimination with contrastive learning which encourages the model to produce invariant representations and align the latent spaces of all the modalities.


More formally, the contrastive loss function for a pair of modalities ($m$,$n$) has the following form:
\begin{equation}
\small    L^{m,n}_{i} =  -log\frac{\operatorname{exp}(\operatorname{sim}(z^m_i,z^n_i)/\tau)}{\sum_{j=1}^{N}\mathds{1}_{[i\neq j]}\operatorname{exp}(\operatorname{sim}(z^m_i,z^n_j)/\tau)} \,,
    \label{eq:contrast}
\end{equation}
where $z$ denotes the embedding after the projection, $i$, $j$ are indices of samples in the current batch of size $N$, $\tau$ is the temperature parameter (scalar), $\mathds{1}_{[k\neq i]} \in \{0,1\}$ is an indicator function evaluating to 1 iff $k \neq i$, and $ sim(u,v) = \frac{ u^{\mathsf{T}}v }{||u|| \  ||v|| } $ denotes the dot product between $\ell_2$-normalized vectors $u$ and $v$ (i.e., cosine similarity).
Eq. \eqref{eq:contrast} is computed across all samples $i$ in the batch, resulting in $L^{m,n} = \sum_{i=1}^N L^{m,n}_{i}$. In addition, we minimize this loss for each possible pairs of modalities. Notice that, since the negatives are drawn from only one modality (see denominator in Eq.~\eqref{eq:contrast}), the loss is asymmetric, \textit{i.e.}, $L^{m,n}$ is not equal to $L^{n,m}$. Therefore, our final loss function (Eq.~\eqref{eq:finalLoss}) includes the loss obtained from all the permutations of two elements drawn with replacement from the set of modalities $\mathcal{M}$:
\begin{equation}
\small     L_{final}= \frac{ \sum_{(m,n) \in \mathcal{M} \times \mathcal{M}}^{ }{\mathds{1}_{[m\neq n]} L^{m,n}}}{|\mathcal{M}|(|\mathcal{M}| - 1)} \,.
    \label{eq:finalLoss}
\end{equation}
Note that, we found empirically that only contrasting different modalities (\textit{i.e.} when $m \neq n$) produces better representations. In addition, we perform temporal augmentations (see Sec.~\ref{subsec:training} for details) to the sequences in order to avoid overfitting and improve performance.

\subsection{Discrete Emotion Recognition}
\label{subsec:emotionRec}
Following the common practice \cite{chen2020simple,qian2021spatiotemporal}, in order to perform the downstream task (i.e., discrete emotion recognition),
we discard the projection layers (described in Sec.~\ref{subsec:unsupervised}) and use the 512-dimensional feature representation extracted from each backbone. The extracted features are concatenated (e.g., for 3 modalities, the combined vector holds 3$\times$512 number of features) and given to a prediction layer, that shares the same design with the projection heads (i.e., $fc$+RELU+$fc$) where its output is the emotion classes. The aforementioned prediction layers are trained with the emotion labels using the cross entropy loss and a variant of it (see Sec.~\ref{subsec:ablation} for details).

\subsection{Implementation Details}
\label{subsec:training}
The training is performed with the SGD optimizer with the momentum of 0.9 and the weight decay of 0.001. All models are trained with the batch size of 32 (or 64) while the batch size of our downstream task is 64 (or 128). The learning rate is initialized as 0.001. We create a linear scheduler to vary the learning rate over the training process such that at every 5 epochs for CMU-MOSEI \cite{Zadeh2018MultimodalLA} and every 100 for RAVDESS \cite{Livingstone2018}, we multiply the learning rate with 0.9 (notice that RAVDESS dataset is much smaller than CMU-MOSEI).
We do not apply any ``spatial'' data augmentation (e.g., random crops, blurs or color distortions), but data sampling can have overlapping sequences. For example, a video segment from $t$ to $t+10$, and another video segment from $t+5$ to $t+15$ can be used in the same training. This is referred as augmentation in the temporal dimension.
We set the number of epochs to 2000, but we also define a \emph{patience parameter} such that: if after 100 consecutive epochs the validation performance does not change, then we stop the training. In practice, the maximum number of epochs was never been reached because the patience parameter stopped the training before. The temperature scalar $\tau$ is taken as 0.07.

\begin{table}[t]
\centering
\caption{Results of the proposed and baseline methods on RAVDESS dataset \cite{Livingstone2018} in terms of accuracy (ACC).}
\label{tab_baseline}
\resizebox{0.98\linewidth}{!}
{
\rowcolors{2}{white}{lightblue}
\begin{tabular}{@{}lccccccccccccccccc@{}}
\toprule

Methods & Actor & Facial  & Acoustics & Facial & ACC  \\ 
  & Split & Images &  & Landmarks & (\%)  \\ 
 \midrule
Unimodal & \cmark & \cmark &  &  & 60.80 \\ 
Unimodal & \cmark &  & \cmark &  & 58.50 \\ 
Unimodal & \cmark &  &  & \cmark & 62.05  \\ 
Late Fusion & \cmark & \cmark & \cmark & \cmark & 64.10 \\ 
Attention Mec. & \cmark & \cmark & \cmark & \cmark & 65.40 \\ 
\textbf{Ours} & \cmark & \cmark & \cmark &  & 63.78 \\ 
\textbf{Ours} & \cmark &  & \cmark & \cmark & \textbf{77.10} \\ 
\textbf{Ours} & \cmark & \cmark & \cmark & \cmark & \textbf{78.54} \\ 
\midrule
Unimodal &  & \cmark &  &  & 72.80  \\ 
Unimodal &  &  & \cmark &  & 75.90 \\ 
Unimodal &  &  &  & \cmark & 76.35 \\ 
Late Fusion &  & \cmark & \cmark & \cmark & 80.72 \\ 
Attention Mec. &  & \cmark & \cmark & \cmark & 81.80 \\ 
\textbf{Ours} &  & \cmark & \cmark & & 80.32 \\ 
\textbf{Ours} &  & & \cmark & \cmark & \textbf{89.50} \\ 
\textbf{Ours} &  & \cmark & \cmark & \cmark & \textbf{93.17} \\ 
\bottomrule 
\end{tabular}
}
\end{table}

\begin{table*}[t]
\centering
\caption{
Results of the proposed and the baseline methods on CMU-MOSEI \cite{Zadeh2018MultimodalLA} in terms of weighted accuracy (\textit{w}-ACC) and F1 measure. wout/ text stands for the experiments when the text modality is not used while all other modalities are used.} 
\label{tab_baseline2} 
\resizebox{1.0\linewidth}{!}
{\small
\rowcolors{2}{white}{lightblue}
\begin{tabular}{@{}lccccccccccccccccc@{}}
\toprule
Methods & \multicolumn{2}{c}{Happy} & \multicolumn{2}{c}{Sad} & \multicolumn{2}{c}{Anger} & \multicolumn{2}{c}{Surprise} & \multicolumn{2}{c}{Disgust} & \multicolumn{2}{c}{Fear} & \multicolumn{2}{c}{Overall} \\ 
 \cmidrule(lr){2-3} \cmidrule(lr){4-5} \cmidrule(lr){6-7} \cmidrule(lr){8-9} \cmidrule(lr){10-11} \cmidrule(lr){12-13} \cmidrule(lr){14-15}
 & \textit{w}-ACC &  F1  & \textit{w}-ACC &  F1 & \textit{w}-ACC &  F1 & \textit{w}-ACC &  F1 & \textit{w}-ACC &  F1 & \textit{w}-ACC &  F1 & \textit{w}-ACC &  F1  \\ 
 \midrule
Late Fusion & 59.71 &	60.17 &	54.17 &	27.97 &	54.58 &	34.58 &	50.01 &	3.31 &	54.29 &	34.10 &	54.92 &	22.83 &	54.60 & 30.50 \\

Attention Mec. & 61.27 & 61.61 & 55.80 & 36.09 & 54.92 & 37.06 & 50.34 & 	5.66 & 55.84 & 44.15 & 57.25 & 43.71 & 55.90 & 38.00 \\

\textbf{Ours} wout/ text & {63.96} & {61.84} & {50.71} &	{12.41} & {54.88} & {26.59} & {50.30} & {2.76} & {58.37} & {35.44} & {54.79} & {27.56} & {55.50} & {27.77} \\

\textbf{Ours} & \textbf{68.82} & \textbf{69.20} & \textbf{62.93} &	\textbf{55.70} & \textbf{67.91} & \textbf{70.09} & \textbf{62.93} & \textbf{72.73} & \textbf{72.91} & \textbf{74.25} & \textbf{64.49} & \textbf{74.85} & \textbf{66.70} & \textbf{69.50} \\

\bottomrule 
\end{tabular}
}
\end{table*}

\section{Experiments and Results}
\label{sec:experiments}

\subsection{Datasets and Evaluation Metrics}
\label{subsec:datasets}

We used the speech part of \textbf{RAVDESS} dataset \cite{Livingstone2018}, containing 2880 audio-visual recordings acted by 24 \emph{professional actors} 
pronouncing two lexically identical statements.
Each recording was labeled in terms of one of the eight categorical emotions (anger, happiness, disgust, fear, surprise, sadness, calmness and neutral), while the emotions were expressed with two intensity (normal or strong). RAVDESS is class-balanced except the neutral class, which was elicited 50\% less time than the other emotion classes. 
We adapted two cross-validation settings following the methods \cite{GhalebPopaAl19,GhalebNiehuesAl20,RadoiBirhalaAl21,SongCaiAl21,ShirianTripathiAl21,BhavanChauhanAl19,BeardDasAl18,JaratrotkamjornChoksuriwong19,AbdullahAhmadAl20,TiwariRathodAl21}. The first setting considers the identities of the actors such that the training (validation) and the corresponding testing \textit{k}-folds have no overlap in terms of actors (shown as \emph{actor-split=\cmark} hereafter). The second setting, instead, applies standard \textit{k}-fold cross-validation (i.e., \emph{actor-split=\xmark}). In both settings, \textit{k} was taken as 10 and the reported results are in terms of accuracy (ACC), which is averaged over the 10-folds, supplying fair comparisons with the MER SOTA \cite{GhalebPopaAl19,GhalebNiehuesAl20,RadoiBirhalaAl21,SongCaiAl21,BeardDasAl18,JaratrotkamjornChoksuriwong19,TiwariRathodAl21}.
As the same statements are being repeated by the actors in {RAVDESS} dataset \cite{Livingstone2018}, the proposed method (as well as the SOTA) are based only on visual and acoustic modalities.

\begin{table}[t]
\centering
\caption{
Results of the proposed method and the SOTA MER methods tested on RAVDESS \cite{Livingstone2018}. ATT stands for attention mechanism.}
\label{tab_ravdess} 
\resizebox{1.0\linewidth}{!}
{\small
\rowcolors{2}{white}{lightblue}
\begin{tabular}{@{}lccccccccccccccccc@{}}
\toprule
Methods & Actor Split & Feature Learning & ACC ($\%$)  \\ 
 \midrule
Human performance \cite{Livingstone2018} & - & - & 80.00 \\ \hline
Ghaleb et al. \cite{GhalebPopaAl19} & \cmark &  \textcolor{blue}{Supervised} & 67.70 \\
Ghaleb et al. \cite{GhalebNiehuesAl20} & \cmark & \textcolor{blue}{Supervised} & 69.40 \\
Ghaleb et al. \cite{GhalebNiehuesAl20} (w/ATT)  & \cmark & \textcolor{blue}{Supervised} & 76.30 \\
Radoi et al. \cite{RadoiBirhalaAl21} & \cmark & \textcolor{blue}{Supervised} & 78.70\\
\textbf{Ours} & \cmark & \textcolor{red}{Unsupervised} &  78.54 \\
 \midrule
Beard et al. \cite{BeardDasAl18} &  & \textcolor{blue}{Supervised} & 58.30 \\ 
Song et al. \cite{SongCaiAl21}  & & \textcolor{blue}{Supervised} & 90.00 \\ 
Tiwari et al. \cite{TiwariRathodAl21} &  & \textcolor{blue}{Supervised} & 93.30 \\ 
\textbf{Ours} &  & \textcolor{red}{Unsupervised} & 93.17 \\
\bottomrule 
\end{tabular}
}
\end{table}

The \textbf{CMU-MOSEI} \cite{Zadeh2018MultimodalLA} is the largest multimodal in-the-wild dataset in the MER domain. It consists of more than 23K utterances, belonging to more than 1000 speakers, collected from YouTube videos. Each utterance is labeled with six emotions: happiness, sadness, anger, fear, disgust, and surprise with a [0,3] Likert scale for the presence of each emotion class. 
Following \cite{Zadeh2018MultimodalLA,DaiCahyawijayaAl21,DelbrouckTitsAl20,MittalBhattacharyaAl20,ShenoySardanaAl20,ChauhanAkhtarAl19,TsaiMaAl20,ZhangLiangqingAl19,WenYouAl21,HoYangAl20}, the emotions were treated as either present or not present (i.e., binary classification), while more than one emotion can be present at the same time, making the task a multi-label problem.
There exist ($\approx$~3000) not-correctly aligned sequences across the modalities. As our approach does not require strict data alignment, we used all sequences as supplied in CMU-MOSEI SDK \cite{CMU_MOSEI_SDK}. In other words, we did not apply any data cleaning, e.g., as in \cite{DaiCahyawijayaAl21}.
We also used the recommended dataset split and the evaluation metrics in \cite{Zadeh2018MultimodalLA}, namely weighted accuracy \cite{Tong2017} (\textit{w}-ACC) and F1-measure.

\begin{table*}[t]
\centering
\caption{
Performance comparisons among the proposed method and the SOTA MER methods tested on CMU-MOSEI \cite{Zadeh2018MultimodalLA} dataset. The results that our method surpasses are given in \colorbox{arylideyellow}{yellow}.}
 \label{tab_mosei} 
\resizebox{1.0\linewidth}{!}
{\small
\rowcolors{2}{white}{lightblue}
\begin{tabular}{@{}lccccccccccccccccc@{}}
\toprule
Methods & \multicolumn{2}{c}{Happy} & \multicolumn{2}{c}{Sad} & \multicolumn{2}{c}{Anger} & \multicolumn{2}{c}{Surprise} & \multicolumn{2}{c}{Disgust} & \multicolumn{2}{c}{Fear} & \multicolumn{2}{c}{Overall} \\ 
 \cmidrule(lr){2-3} \cmidrule(lr){4-5} \cmidrule(lr){6-7} \cmidrule(lr){8-9} \cmidrule(lr){10-11} \cmidrule(lr){12-13} \cmidrule(lr){14-15}
 & \textit{w}-ACC &  F1  & \textit{w}-ACC &  F1 & \textit{w}-ACC &  F1 & \textit{w}-ACC &  F1 & \textit{w}-ACC &  F1 & \textit{w}-ACC &  F1 & \textit{w}-ACC &  F1  \\ 
 \midrule
 
 \multicolumn{15}{c}{\textcolor{red}{Unsupervised Feature Learning Methods}} \\ 
CAE-LR \cite{Koromilas2021UnsupervisedML} &  
\colorbox{arylideyellow}{64.70} & \colorbox{arylideyellow}{65.60} & \colorbox{arylideyellow}{53.20} & \colorbox{arylideyellow}{55.60} & \colorbox{arylideyellow}{61.80} & \colorbox{arylideyellow}{61.90} & \colorbox{arylideyellow}{57.10} & \colorbox{arylideyellow}{70.70} & \colorbox{arylideyellow}{69.00} & \colorbox{arylideyellow}{70.10} & \colorbox{arylideyellow}{60.40} & \colorbox{arylideyellow}{69.20} & \colorbox{arylideyellow}{61.03} & \colorbox{arylideyellow}{65.52} \\
\textbf{Ours} & 68.82 & 69.20 & 62.93 & 55.70 & 67.91 & 70.09 & 62.93 & 72.73 & 72.91 & 74.25 & 64.49 & 74.85 & 66.70 & 69.50 \\
 
 \midrule
 
 \multicolumn{15}{c}{\textcolor{blue}{Fully Supervised Methods}} \\

MESM \cite{DaiCahyawijayaAl21} & \colorbox{arylideyellow}{64.10} & 72.30 & 63.00 & \colorbox{arylideyellow}{46.60} & \colorbox{arylideyellow}{66.80} & \colorbox{arylideyellow}{49.30} & 65.70 & \colorbox{arylideyellow}{27.20} & 75.60 & \colorbox{arylideyellow}{56.40} & 65.80 & \colorbox{arylideyellow}{28.90} & 66.80 & \colorbox{arylideyellow}{46.80} \\

Zhang et al. \cite{ZhangLiangqingAl19} & 71.70 & -- & 64.30 & -- & \colorbox{arylideyellow}{66.60} & -- &	\colorbox{arylideyellow}{62.30} & -- & \colorbox{arylideyellow}{72.50}	& -- & 64.60 & -- & 67.00 & --  \\

FE2E \cite{DaiCahyawijayaAl21} & \colorbox{arylideyellow}{65.40} & 72.60 & 65.20 & \colorbox{arylideyellow}{49.00} & \colorbox{arylideyellow}{67.00} & \colorbox{arylideyellow}{49.60} & 66.70 & \colorbox{arylideyellow}{29.10} & 77.70 & \colorbox{arylideyellow}{57.10} & \colorbox{arylideyellow}{63.80} & \colorbox{arylideyellow}{26.80} & 67.60 & \colorbox{arylideyellow}{47.40} \\

Graph-MFN \cite{Zadeh2018MultimodalLA} & \colorbox{arylideyellow}{66.30} & \colorbox{arylideyellow}{66.30} & \colorbox{arylideyellow}{60.40}	& 66.90 & \colorbox{arylideyellow}{62.60} & 72.80 & \colorbox{arylideyellow}{53.70} & 85.50 & \colorbox{arylideyellow}{69.10} & 76.60 & \colorbox{arylideyellow}{62.00} & 89.90 & \colorbox{arylideyellow}{62.35} & 76.33 \\

Delbrouck et al. \cite{DelbrouckTitsAl20} & -- & \colorbox{arylideyellow}{64.00}	& -- & 67.90	& -- & 74.70 & -- & 86.10 & -- & 83.60 & -- & 84.00	& -- & 76.72 \\

Huynh et al. \cite{HuynhVan2021} & \colorbox{arylideyellow}{62.70} & \colorbox{arylideyellow}{63.00} & \colorbox{arylideyellow}{54.40} & 69.70 & \colorbox{arylideyellow}{59.60} & 74.30 & \colorbox{arylideyellow}{50.60} & 85.70 & \colorbox{arylideyellow}{66.00} & 81.30 & \colorbox{arylideyellow}{52.90} & 86.40 & \colorbox{arylideyellow}{57.70} & 76.73 \\

Khare et al. \cite{KhareParthasarathyAl21} & \colorbox{arylideyellow}{68.10} & \colorbox{arylideyellow}{68.20} & 64.30 & 72.40 & \colorbox{arylideyellow}{67.30} & 74.80 & 65.10 & 87.70 & 73.60 & 82.40 & \colorbox{arylideyellow}{63.00} & 86.60 & 66.90 & 78.68 \\


CIA \cite{ChauhanAkhtarAl19} & \colorbox{arylideyellow}{51.90} & 71.30 & \colorbox{arylideyellow}{61.80} & 72.90 & \colorbox{arylideyellow}{67.40} & 74.70 & \colorbox{arylideyellow}{58.20} & 86.00 & 74.10 & 81.80 & \colorbox{arylideyellow}{63.90} & 87.80 & \colorbox{arylideyellow}{62.88} & 79.08 \\

Tsai et al. \cite{TsaiMaAl20} & 71.00 & 71.00 & 75.00 & 72.10 & 78.30 & 75.00 & 90.50 & 86.10 & 83.00 & 82.50 & 91.70 & 87.80 & 81.58 & 79.08 \\

Wen et al. \cite{WenYouAl21} & 72.50 & 72.60 & 75.60 & 70.70 & 77.10 & 74.90 & 90.60 & 86.10 & 85.00 & 83.20 & 91.70 & 87.80 & 82.08 & 79.22 \\


Shenoy et al. \cite{ShenoySardanaAl20} & 70.00 & \colorbox{arylideyellow}{68.40} & 76.10 & 74.50 & 83.10 & 80.90 & 87.40 & 84.00 & 90.30 & 87.30 & 89.70 & 87.00 & 82.77 & 80.35 \\


M3ER \cite{MittalBhattacharyaAl20}	& -- & 	78.00	& -- & 	87.30	& -- & 	81.60	& -- & 93.20	& -- & 84.40		& -- & 91.80		& -- & 86.05 \\

\bottomrule 
\end{tabular}
}
\end{table*}

\subsection{Comparisons with the Baseline Methods}
\label{subsec:ablation}


We compare the proposed approach with the following baseline methods. These baselines are all \emph{supervised} such that cross-entropy and binary cross-entropy losses were used for RAVDESS \cite{Livingstone2018} and CMU-MOSEI \cite{Zadeh2018MultimodalLA}, respectively.
The corresponding results are given in Tables~\ref{tab_baseline} and ~\ref{tab_baseline2}. 
\\
\noindent
\textbf{Unimodal Learning.} Each modality 
was trained with its associated backbone (described in Sec.~\ref{subsec:features}) followed by two fully connected ($fc$) layers with a ReLU activation function. 
The best results were obtained with the following parameter settings.
For acoustic data, the learning rate was initialized with 0.001 and decreased by multiplying it with 0.9 at every 10 epochs. The batch size was 32 and number of epoch was 100. For facial images, the learning rate was 0.01, number of epoch was 150 and the momentum was 0.9. For facial landmarks, the learning rate was 0.001, momentum was 0.9 and the number of epochs was set as the proposed method with patience parameter.
\\
\noindent
\textbf{Late Fusion.} Recall that late fusion was applied by several SOTA methods, e.g., \cite{RadoiBirhalaAl21,SongCaiAl21,ShenoySardanaAl20}. Given the modalities and the backbones described, we concatenated the feature embeddings of each modality, and fed them to a shallow network composed of two $fc$ layers with a ReLU activation function. 
The batch size was taken as 32, the number of epochs was set by the patience parameter, the learning rate and momentum were taken as 0.001 and 0.9, respectively. \\
\noindent
\textbf{Attention Mechanism.} As mention in Sec.~\ref{sec:relatedwork}, attention mechanism has been frequently applied in MER, hence we adapted it as a baseline too. We first concatenated the feature embeddings obtained from each modality (512 features extracted from each backbone as in our method) and then applied the multi-head attention mechanism of \cite{Vaswani2017}. The batch size was 64, the learning rate was 0.001, and the number of epochs was set to 2000 with the patience parameter described in Sec.~\ref{subsec:training}.
The same scheduler as the proposed method was used.

As seen in Table~\ref{tab_baseline}, our unsupervised feature learning method outperforms all of the supervised baselines when acoustic and facial landmarks are involved. It is notable that, in the visual domain, the facial landmarks are more effective than the facial images.
Out of all baseline methods, late fusion and attention mechanism surpass the unimodal setups, while attention mechanism achieves slightly better results than the late fusion. Overall, all methods perform better in the \emph{actor-split=\xmark} setting compared to their \emph{actor-split=\cmark} counterpart. This is perhaps as a result of having more training data in the \emph{actor-split=\xmark} setting.
With reference to Table \ref{tab_baseline}, we have further investigated the contribution of used modalities with respect to different emotions by inspecting the confusion matrices. Our observation is that there is no particular modality or a pair of modality which performs better for a specific emotion class(es). 

Given the better performances of late fusion and attention mechanism compared to unimodal learning in Table~\ref{tab_baseline}, we inherited them to test on CMU-MOSEI dataset \cite{Zadeh2018MultimodalLA} when four modalities (text, facial images, acoustic and facial landmarks) are used. Additionally, in order to investigate the contribution of the \emph{text modality}, we compare the results of the proposed method 
with the performance of the proposed method when the text is discarded (shown as wout/ text). The corresponding results can be seen in Table~\ref{tab_baseline2}. 
Our method outperforms the baselines for all emotion classes (especially for surprise) as well as on average (see Table~\ref{tab_baseline2}). Also, the performances of our method do not fluctuate across different emotion classes, meaning that our method generalize better than the baseline methods. 
In overall there exist a drop of 11.2\% and 41.73\% for w-ACC and F1-measure, respectively, when the text modality is discarded from the pipeline of the proposed method, showing the positive contribution of the text modality. 

\subsection{Comparisons with the State-of-the-art Methods}
\label{subsec:SOTA}



We compare our approach with several SOTA MER methods. Concerning RAVDESS \cite{Livingstone2018}, the performances are given in Table~\ref{tab_ravdess}. The fact that ``human performance'' is not 100$\%$ presents the difficulty of MER task. It is remarkable that our approach surpasses several supervised competitors: \cite{GhalebPopaAl19,GhalebNiehuesAl20,BeardDasAl18,SongCaiAl21} with a margin of 2-35\% despite working in a more difficult (unsupervised) setting. It also performs on par with supervised approaches:  \cite{RadoiBirhalaAl21,TiwariRathodAl21}.  
The results for CMU-MOSEI \cite{Zadeh2018MultimodalLA} are given in Table~\ref{tab_mosei}.
There exist a very recent unsupervised feature learning approach (namely CAE-LR \cite{Koromilas2021UnsupervisedML}) tested on CMU-MOSEI \cite{Zadeh2018MultimodalLA} for multimodal sentiment analysis. CAE-LR \cite{Koromilas2021UnsupervisedML} achieved the best results for multimodal sentiment analysis compared to other unsupervised counterparts.
Motivated by this,
we adapted the authors' code 
for MER. Instead of applying Logistic Regression, 
we performed Linear Evaluation \cite{UHAR_BMVC2021}, which is the common protocol for unsupervised learning if the downstream task is classification (notice that we apply it for the proposed method as well, i.e., the prediction layer). 
For all emotion classes and on overall, our method achieves much better results than CAE-LR \cite{Koromilas2021UnsupervisedML}, showing the effectiveness of the contrastive loss in multimodal setting compared to convolutional autoencoders.
It is worth noting that, on average, our method is better than several fully supervised techniques: MESM \cite{DaiCahyawijayaAl21}, FE2E \cite{DaiCahyawijayaAl21}, Graph-MFN \cite{Zadeh2018MultimodalLA}, \cite{HuynhVan2021}, CIA \cite{ChauhanAkhtarAl19}. Considering that these methods integrate relatively complex supervised techniques; attention mechanisms, transformers, graphs, the better performance of our method is very promising. 


\section{Conclusion}
\label{sec:conclusion}
We presented an unsupervised multimodal feature learning approach, which was tested on discrete emotion recognition. Our method is a pioneer in the MER litreature, being based on pairwise contrastive learning. Experiments show that the performance
of our approach is better than the supervised baselines and unsupervised counterpart, while being competitive to several complex supervised SOTA and even surpassing a few. Being an unsupervised feature learning method, the proposed approach is transferable to other domains without retraining (not even tuning) the representation model itself. 


The proposed method keeps the modality pairings the same for all data (i.e., emotions) and the way we learn the features gives equal importance to each modality. An alternative could be having different modality pairings for different emotion classes. This will be further investigated as future work. 

\section*{Acknowledgment}
This work was supported by the EU H2020 SPRING project (No. 871245) and by Fondazione VRT.

\bibliographystyle{IEEEtran}
\bibliography{mybib}

\end{document}